# Machine Learning and Deep Learning Models for Short Term Electricity Price Forecasting in Australia's National Electricity Market

Wei Lu [1], Jay Wang [1]*, Dingli Duan [2], Ding Mao [3], Caiyi Song [4], John Huang [4]

[1] School of Engineering, Computer & Mathematical Sciences, Auckland University of Technology (AUT), Auckland 1010, New Zealand

[2] Department of Thermal Energy and Power Engineering, Yantai University, Yantai, China

[3] Department of Building Environment and Equipment, School of Civil Engineering, Hefei University of Technology, Hefei, China

[4] Green Gold Energy Pty Ltd, Adelaide SA 5063, Australia

**Abstract:** Short term electricity price forecast is essential in competitive power markets, yet electricity price series exhibit high volatility, irregularity, and non-stationarity. This phenomenon is pronounced in the South Australian region of the National Electricity Market, where high renewable penetration drives price volatility and frequent negative price intervals, while structural changes such as the transition to five-minute settlement further complicate forecast. To address these challenges, this study develops a unified benchmark framework. Under identical data preprocessing, feature engineering with lag features, rolling statistics, cyclic temporal encodings, and so on, and an 85% to 15% chronological train test split, six algorithms are systematically compared, including AWMLSTM, CatBoost, GBRT, LSTM, LightGBM, and SVR. The results show that for price prediction, tree-based models, especially GBRT with an R squared value of 0.88, generally outperform LSTM and SVR. However, all models achieve a mean absolute percentage error above 90%, and more than 65% of GBRT predictions have relative errors above 10%, which highlights the inherent difficulty of price forecast. For demand prediction, all models perform substantially better than in price prediction. AWMLSTM and GBRT achieve an $R^2$ value of 0.96 with mean absolute percentage error below 32%, and GBRT has 74.37% of samples within 5% error, while LSTM and SVR perform less accurately in both tasks. Future improvements should focus on hybrid models such as tree plus transformers, data augmentation for extreme events, and error correction to better capture price spikes.

*Corresponding author: Dr. Jay Wang

Email address: jay.wang@aut.ac.nz

Phone number: +64 9 921 9001 ext. 33206

Address: Auckland University of Technology, ECMS (C-46), WZ Building, 6 St Paul Street, Auckland 1010, New Zealand

**Nomenclature**

| | |
|---|---|
| AEMO | Australian energy market operator |
| ARIMA | auto-regressive integrated moving average |
| AIC | Akaike information criterion |
| AWMLSTM | attention-based long short-term memory variant |
| CatBoost | categorical boosting |
| CRPS | continuous ranked probability score |
| EFB | exclusive feature bundling |
| EPF | electricity price forecasting |
| GBRT | gradient boosting regression trees |
| GOSS | gradient-based one-side sampling |
| GPR | gaussian process regression |
| ISO-NE | the independent system operator New England |
| LSTM | long short-term memory |
| LightGBM | light gradient boosting machine |
| MAE | mean absolute error |
| MAPE | mean absolute percentage error |
| NEM | national electricity market |
| mTCN | multiple temporal convolutional network |
| RMSLE | root mean squared logarithmic error |
| RMSE | root mean square error |
| $R^2$ | coefficient of determination |
| SARIMAX | seasonal autoregressive integrated moving average with exogenous factors |
| SVR | support vector regression |
| WRMSSE | weighted root mean squared scaled error |
| XGBoost | extreme gradient boosting |

# 1. Introduction

Reliable short-term electricity price forecasting (EPF) is indispensable in competitive power markets because it informs bidding, dispatch, hedging, and the scheduling of flexible resources [1, 2]. Wholesale electricity prices, however, are formed in a system that must continuously balance supply and demand in real time while operating with limited large-scale storage capacity [3]. These physical and economic conditions generate price series that are highly volatile and often non-stationary, which is why EPF has remained a central topic in both power system economics and energy forecasting research [3, 4].

Over the past two decades, the methodological landscape of EPF has broadened from statistical benchmarks to machine learning, deep learning, and probabilistic forecasting [3, 5]. Early work was dominated by auto-regressive models or auto-regressive integrated moving average (ARIMA) models, which established influential baselines for short-term prediction [3, 6]. Recent studies showed that machine learning and deep learning methods can capture nonlinear relationships and richer temporal dependencies, and empirical comparisons confirmed that their relative performance depends strongly on data design and implementation choices [7, 8]. In parallel, probabilistic and distributional forecasting gained prominence because market participants require not only point estimates but also measures of uncertainty and tail risk [5, 9]. Oriented benchmark reviews further emphasis that methodological progress in EPF should be assessed with transparent evaluation protocols and operationally relevant performance criteria [8, 10].

These issues are especially salient in the Australian National Electricity Market (NEM), and particularly in South Australia. Rapid growth in rooftop photovoltaic capacity and variable renewable generation has reshaped effective demand profiles and wholesale price behaviour across the ANEM [11, 12]. South Australia is a particularly demanding forecasting environment because high wind penetration has been associated with downward pressure on prices and a greater incidence of negative-price intervals [13, 14]. Recent evidence also identifies South Australia as one of the most volatile electricity price environments in the NEM [15]. Widespread studies in Australia further indicate that wind and utility scale solar generation exert a measurable merit order effect on wholesale electricity prices, which implies that renewable output should be treated as a core explanatory input in multivariate forecasting models [16].

The forecast in the NEM is further complicated by structural and regulatory changes that weaken the stability of historical relationships. The transition from 30-minute to 5-minute settlement increased the operational relevance of short-interval price signals, while the Hazelwood exit introduced a clear structural disruption into Australian price dynamics [17, 18]. More generally, model comparisons can be misleading when studies rely on different sampling frequencies, forecast horizons, prepossessing choices, and validation protocols [10, 19]. These concerns are especially important in EPF, where transformations, variable selection, and the representation of intraday dependence can materially affect both estimation and reported performance [20, 21]. Recent benchmark-oriented studies therefore call for stronger comparative discipline, transparent baselines, and more reproducible evaluation designs [8, 19].

The present study develops a unified benchmarking framework for short-term EPF in South Australia. The empirical design benchmarks six specific algorithms on a common footing. The classical machine-learning baselines are support vector regression (SVR) [22] and gradient boosting regression trees (GBRT) [23]. The tree-based boosting models are light gradient boosting machine (LightGBM) [24] and categorical boosting (CatBoost) [25]. The deep learning benchmark includes long short-term memory (LSTM) [26] and an attention-based long short-term memory variant (AWMLSTM) [27]. To make the comparison fair, all models use the same covariate set, the same preprocessing strategy, and the same validation and test sets for performance assessment [10, 19]. The purpose is to provide directly comparable evidence on forecasting performance in a renewable-intensive and high-volatility regional market, while also examining behaviour in operationally important regimes such as price spikes and negative-price intervals [15, 28]. In this way, the study responds to recent calls for harmonised evaluation and more robust comparisons across heterogeneous model classes [8, 19].

## 2. Literature Review

Electricity price forecasting research is commonly grouped into statistical or econometric models, classical machine learning methods, deep learning architectures, and probabilistic approaches. Wang et al. applied the seasonal autoregressive integrated moving average with exogenous factors (SARIMAX) (1,1,2)×(3,1,0,7) model with exogenous prices, internal flows, and external flows to forecast electricity prices in the German-Luxemburg bidding zone, which achieved an Akaike information criterion (AIC) of 9853 and a mean absolute percentage error of 16.94% [29]. Abroun et. al. found that SVR captures trends and peaks in monthly electricity prices but lacks accuracy, and thus proposed modified SVR, which adjusts the first three predicted values and proves efficient for long-term monthly forecasting, though the mean average percentage error remains above 5% [30]. Hu et al. adopted an fast Fourier transform (FFT)-based hybrid algorithm to achieve accurate medium-to-long term electricity price forecasting, delivering average accuracy improvements of 57.21% and 49.69% over long short-term memory and transformer respectively, while reducing data requirements and forecasting volatility [31]. These three studies illustrate a range of forecasting approaches, from traditional econometric models to advanced hybrid deep learning methods, and each contributes distinct strengths and reveals persistent challenges in accuracy, data efficiency, and forecast stability.

Within the machine learning set used in this study, the alignment between selected algorithms and the literature is evident. Das et al. employed and applied an ensemble model combining Gaussian Process Regression (GPR) and SVR for electricity price forecasting in the German power market, which achieved mean absolute percentage error (MAPE) values of 6.579, 7.063, and 10.561 for the years 2021, 2022, and 2023, respectively[32]. Kuşkaya & Bilgili evaluated 19 machine learning algorithms for forecasting US electricity prices, finding that CatBoost regressor with time-series strategy achieved the highest accuracy with a mean absolute error

(MAE) of 0.0081 and root mean squared logarithmic error (RMSLE) of 0.0093, while linear models with explicit seasonal features yielded even lower errors of 0.0079 and 0.0091, respectively [33]. Nasios et al. addressed point and probabilistic forecasting by blending gradient boosted trees and neural networks, which achieved a validation weighted root mean squared scaled error (WRMSSE) of 0.549 and a final submission score of 0.552 in the the Makridakis fifth forecasting competition [34]. These studies collectively highlight the effectiveness of diverse machine learning approaches in energy and forecasting competitions, which provided a solid foundation for comparison with deep learning methods.

The deep learning side of the benchmark is likewise anchored to specific references. Bâra et al. analyzed the predictive capability of an LSTM architecture on a dataset of Romanian economic and market variables, and proposed a data selection and parameterization framework that outperforms Extreme Gradient Boosting (XGBoost) with shorter training time, which achieved MAE values from 0.0155 to 0.0509 [35]. Zi et al. found that the TiDE model achieves high short-term accuracy for 5-minute-ahead forecasts, with an coefficient of determination ($R^2$) of 0.952, MAE of 0.150, and root mean square error (RMSE) of 0.349, but exhibits larger errors for longer horizons, such as a 1 day-head forecast with an $R^2$ of 0.712, MAE of 0.507, and RMSE of 0.856, and shows limited responsiveness to sudden weather changes [36]. Yang et al. utilized a short-term price forecasting model based on a multiple temporal convolutional network (mTCN) and attention-long LSTM, which achieves a MAE of 2.96, a MAPE of 0.29, and a RMSE of 4.45 on the independent system operator New England (ISO-NE) dataset [37]. However, probabilistic EPF adds another dimension to this comparison because electricity prices exhibit spikes, asymmetry, and heavy tails that are poorly summarised by a single expected value.

Probabilistic EPF adds another dimension to this comparison because electricity prices exhibit spikes, asymmetry, and heavy tails that are poorly summarised by a single expected value. Cornell et al. advocated a quantile regression ensemble model for probabilistic forecasting in South Australia's volatile NEM region, which achieved median forecasts that significantly outperformed Australian energy market operator (AEMO)'s pre-dispatch predictions, with a mean absolute error of 41.12 and a median absolute error of 17.26 in 2021 [15]. Marcjasz et al. leveraged a distributional neural network that, on German day-ahead electricity market data, outperforms benchmarks by over 7% in continuous ranked probability score and by 8% in per-transaction profits [38]. Berrisch and Ziel introduced a novel method for combining multivariate probabilistic forecasts via smoothing that captures dependencies between quantiles and marginals, extends online continuous ranked probability score (CRPS) learning, and outperforms simpler weighting schemes in day-ahead electricity price forecasting [39]. These studies collectively demonstrate the growing sophistication and effectiveness of probabilistic approaches in electricity price forecasting, yet despite these advances, a key challenge remains in that model rankings often vary with hyperparameter tuning, preprocessing, and calibration-window choices. Some main research method and contents are summarised in **Table 1**.

Table 1. The main research method and contents

| Reference | Research Method | Research Content |
|---|---|---|
| Wang et al. (2022) [29] | SARIMAX $(1,1,2)\times(3,1,0,7)$ with exogenous variables | This model achieved a maximum of likelihood function of –4876, a maximum AIC of 9853, and a minimum MAPE of 16.94%. |
| Nasios et al. (2022) [34] | Blended gradient boosted trees and neural networks | Validation WRMSSE stood at 0.549 and the test score at 0.552, a difference of only 0.003. |
| Yang et al. (2022) [37] | mTCN and attention-long LSTM | The model achieved an MAE of 2.96, a MAPE of 0.29, and an RMSE of 4.45, which all outperformed the other seven comparison models. |
| Bâra et al. (2023) [35] | LSTM | Although LSTM reduced the error compared to XGBoost, the values in 2022 were still two to three times those in 2019. |
| Marcjasz et al. (2023) [38] | Distributional neural network | Over 7% in continuous ranked probability score and increased per transaction profits by 8%. |
| Berrisch & Ziel (2024) [39] | Multivariate probabilistic forecasting with CRPS learning | The algorithm employs adaptive weights across time, quantiles, and marginals, which outperforms simpler weighting schemes such as naive. |
| Cornell et al. (2024) [15] | Quantile regression ensemble (probabilistic EPF) | Predispatch predictions with an MAE of 41.12 and a median absolute error of 17.26 in 2021. |
| Abroun et al. (2024) [30] | Modified SVR | Although the MAPE remains above 5%, the improved model effectively generates long-term monthly forecasts. |
| Zi et al. (2025) [36] | Time-series dense encoder (TiDE) | For 5-minute-ahead forecasts achieved R² 0.952, MAE 0.150, RMSE 0.349; for 1-day-ahead forecasts performance declined to R² 0.712, MAE 0.507, RMSE 0.856; limited responsiveness to sudden weather changes. |
| Hu et al. (2025) [31] | FFT-based hybrid deep learning model | Compared to LSTM and Transformer algorithms, the model improved forecasting accuracy by 57.21% and 49.69%, respectively. |

| Das et al. (2025) [32] | Ensemble model combining GPR and SVR | The model decreases MAE by approximately 29% and 12.2%, respectively. |
| Kuşkaya & Bilgili (2026) [33] | Comparative machine learning study (19 models, including CatBoost) | Linear models that included Ridge and Bayesian Ridge outperformed tree-based models, with MAE at 0.0079 and RMSLE at 0.0091, which highlights the importance of explicit seasonal features for linear methods. |

A key challenge in electricity price forecasting is that model rankings often vary with hyperparameter tuning, preprocessing, and calibration-window choices. There is still limited directly comparable evidence on the six precisely specified algorithms used in this study: SVR, GBRT, LightGBM, CatBoost, LSTM, and AWMLSTM. There is a lack of evidence on their relative performance under a common covariate set, preprocessing pipeline, and validation test design, which is a gap that the literature on ranking instability explicitly identifies. This gap is especially important in South Australia, where benchmark evidence should reflect not only average accuracy but also robustness during price spikes and negative-price intervals. Therefore, this study aims to address this gap by conducting a controlled comparison of these six algorithms in a volatile market setting, providing empirical evidence on their relative strengths and limitations.

## 3. Methodology

This section describes the complete experimental workflow employed in this study. It begins with the preparation of the dataset, which included preprocessing steps, train-test splitting, and an illustration of the resulting data. Subsequently, the six forecasting models are introduced, their mathematical formulations are presented, and the training process and evaluation metrics are detailed. The entire methodology is designed to ensure reproducibility and fair comparison across all algorithms.

### 3. 1 Dataset preparation

The dataset used in this study originates from the AEMO and contains chronologically ordered records of electricity price and demand. To construct a robust input space for the forecasting models, a series of feature engineering and cleaning operations are performed. These operations include the creation of lag features, rolling statistics, time-based cyclic encodings, interaction terms, and the inclusion of auxiliary prediction variables. After eliminating highly correlated features and handling anomalous values, the data are normalized and split into training and test sets while preserving temporal order. The following subsections detail each step.

### 3. 1. 1 Preprocessing Steps

The dataset is loaded interactively via a file dialog supported by CSV. The data comes from AEMO. The data are ordered chronologically, with each row representing one time step. Lag features at intervals of 1, 3, 6, 12, and 24 steps, along with rolling statistics including the moving mean, standard deviation, and minimum over windows of 6, 12, and 24 periods, are constructed for both price and demand. Cyclic time features reflecting hour of day, day of week, and day of month are incorporated, in addition to an interaction term formed as the product of price and demand, as well as statistical window features comprising the mean and standard deviation over the preceding 24 observations.

All engineered features are combined with the original predictors corresponding to the average of the most recent 32 predicted prices, the predicted price value closest to the actual price among the most recent 32 predictions, the average of the most recent 32 predicted demands, and the predicted demand value closest to the actual demand among the most recent 32 predictions. Variables exhibiting a Pearson correlation coefficient exceeding 0.95 are eliminated to mitigate redundancy. The dataset is subsequently cleaned by addressing missing or infinite values and clipping outliers beyond three standard deviations. Finally, all input features and the two target variables, price and demand, are normalized to the range [0, 1] using mapminmax.

### 3. 1. 2 Train-Test Split

To preserve the temporal dependency of the series and avoid look-ahead bias, the dataset is divided chronologically into training and testing subsets. The first 85% of the samples are assigned to the training set, while the remaining 15% are used as the test set. This division is performed only after the feature engineering procedure has been completed but before model training begins. Importantly, normalization is applied using parameters estimated exclusively from the training portion, and these parameters are then used to transform the test data, which ensures that no information from the test set leaks into the training process.

### 3. 1. 3 Dataset Illustration

After preprocessing, the feature matrix typically contains several hundred variables, although the exact dimensionality depends on the correlation-based feature removal stage.

The total number of samples is determined entirely by the size of the input file. In a typical execution, the training subset contains train samples and the test subset contains test samples. The time column remains available as a temporal index and can be used for plotting, visualization, and subsequent error analysis.

### 3. 1. 4 Consistency Across Approaches

In this study, six predictive models are employed for forecasting. Nevertheless, all data preparation procedures described above are intentionally designed to remain model-agnostic. The same cleaned, engineered, and normalized dataset is supplied uniformly to all six algorithms. If additional predictive models are introduced in future work, they can likewise be evaluated using exactly the same pre-processed dataset, thereby ensuring that model comparisons are conducted on a fair and consistent basis.

## 3. 2 Model Implementation

**Figure 1** shows the commonalities and differences among six algorithms, AWMLSTM, CatBoost, GBRT, LSTM, LightGBM, and SVR, in terms of model type, core mechanism, sequence handling, categorical feature support, training speed, and hyperparameter tuning difficulty.

AWMLSTM belongs to the recurrent neural network family as a variant of LSTM. Its core mechanism integrates an attention weighted memory or hidden state with the original LSTM gating mechanism. This model handles sequential data natively. It does not support categorical features directly and requires explicit encoding of such features. Hyperparameter tuning of AWMLSTM presents high difficulty due to the additional attention parameters[27]. LSTM is a standard recurrent neural network. Its core mechanism consists of three gates, namely the forget gate, the input gate, and the output gate, along with a memory cell. LSTM processes sequential data natively. It does not directly support categorical features and requires their encoding before training. Hyperparameter tuning of LSTM also exhibits high difficulty because the optimization involves the number of layers, the number of hidden units, the learning rate, and other parameters [26]. SVR is a support vector machine for regression tasks. Its core mechanism maps data into a high dimensional feature space through a kernel function and applies an epsilon insensitive loss to maximize the margin. SVR does not handle sequential data without manual feature engineering. It does not support categorical features and needs one hot encoding or similar transformations. Hyperparameter tuning of SVR is highly difficult due to the sensitivity of the kernel function, the kernel coefficient gamma, and the epsilon parameter [22].

CatBoost is a gradient boosting tree algorithm with a symmetric tree structure. Its core mechanism employs ordered boosting and target encoding to process categorical features. CatBoost supports categorical features natively without any preprocessing. It does not handle sequential data. Hyperparameter tuning difficulty for CatBoost is moderate because the default parameters are robust across many tasks [25]. GBRT denotes a standard gradient boosting tree. Its core mechanism iteratively adds decision trees to fit the residuals of the previous ensemble. GBRT does not handle sequential data. It does not support categorical features and requires manual encoding such as one hot encoding. Hyperparameter tuning difficulty for GBRT is moderate [23]. LightGBM is also a gradient boosting tree. Its core mechanism uses a histogram based algorithm, gradient based one side sampling (GOSS), and exclusive feature bundling (EFB) to achieve efficient training. LightGBM does not handle sequential data. It supports categorical features after the user declares them, though this support carries some limitations. Hyperparameter tuning difficulty for LightGBM is moderate because although the algorithm has many parameters, the literature provides clear guidelines for their adjustment [24].

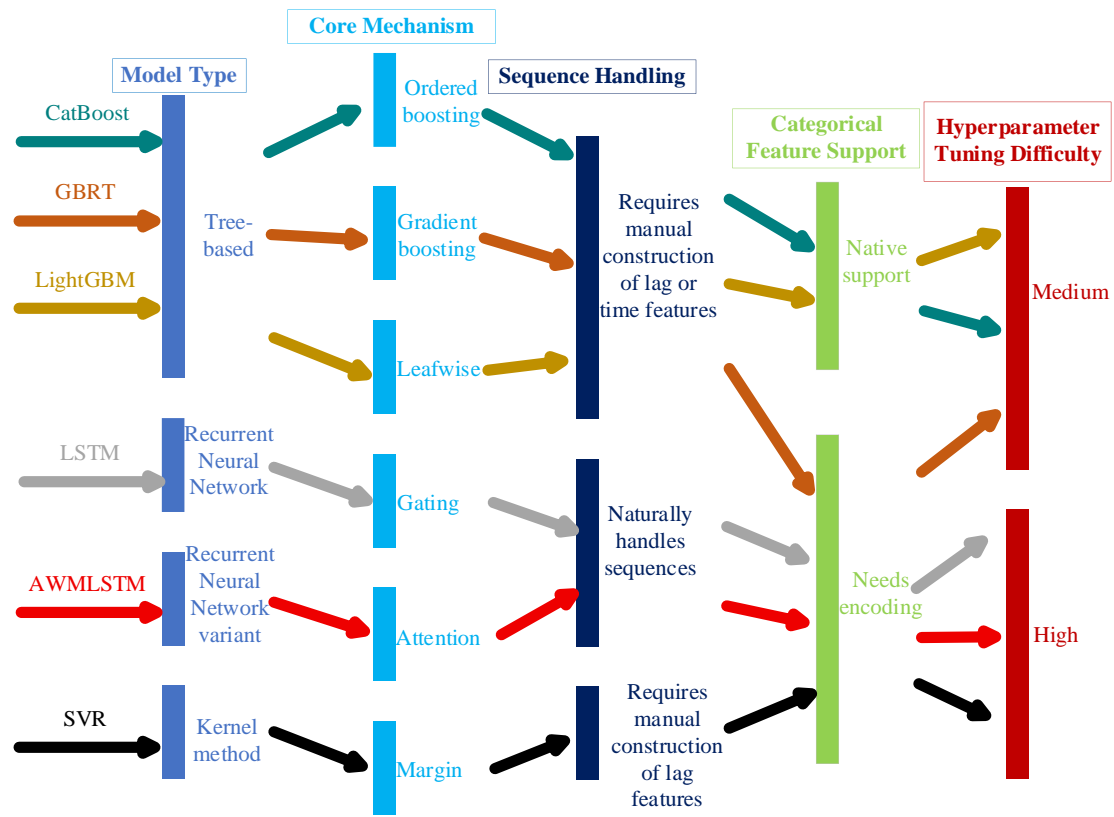

**Figure 1.** Algorithm difference comparison

**AWMLSTM**

AWMLSTM enhances LSTM with an attention mechanism that computes a weighted sum of all hidden states, which allows the model to focus on the most relevant time steps [40].

Attention mechanism over hidden states $\{h_i,...h_T\}$:

$$e_t = v^T \tanh(W_h h_t + b) \quad (1)$$

$$\alpha_t = softmax(e_t) \quad (2)$$

$$c = \sum_{t=1}^{T} \alpha_t h_t \quad (3)$$

$$\hat{y} = g([c; h_T]) \quad (4)$$

Where, $e_t$ is the attention score for hidden state; $v$, $W_h$, and $b$ are the learnable parameters; $\alpha_t$ is the attention weight (normalized score) for $h_t$; $c$ is the context, the weighted sum of all hidden states; $h_T$ is the last hidden state; $[c; h_T]$ is the concatenation of context vector and last hidden state; $g$ is the output layer (e.g., dense layer) producing the final prediction $\hat{y}$.

**CatBoost**

Categorical Boosting (CatBoost) is a gradient boosting framework optimized for datasets with categorical features. It uses ordered target statistics for categorical encoding and ordered boosting to avoid target leakage and prediction shift[25].

Additive model update:

$$F_m(x) = F_{m-1}(x) + \eta h_m(x) \quad (5)$$

where, $\eta$ is the learning rate, $h_m(x)$ is the tree trained on the gradient of the loss function.

Regularized objective:

$$Obj = \sum_{i=1}^{N} L(y_i, F(x_i)) + \sum_{m=1}^{M} \Omega(h_m) \quad (6)$$

where, $\Omega(h_m)$ is the regularization term penalizing the complexity of tree $h_m$, $M$ is the total number of trees, $N$ represents the total number of training samples.

**GBRT**

Gradient Boosted Regression Trees (GBRT) is an ensemble method that builds a strong predictor by adding weak learners (typically shallow regression trees) sequentially. Each new tree fits the negative gradient (pseudo-residuals) of a loss function to progressively reduce prediction error [23].

Additive model update:

$$F_m(x)=F_{m-1}(x)-\gamma_m h_m(x) \tag{7}$$

where, $F_m(x)$ is the model prediction at step $m$, $F_{m-1}(x)$ is the model prediction from the previous step, $h_m(x)$ is the regression tree added at step $m$, fitted to pseudo-residuals, $\gamma_m$ is the step size (learning rate) that scales the contribution of $h_m(x)$.

Objective (empirical risk minimization):

$$Obj=\sum_{i=1}^{N} L(y_i, F(x_i)) \tag{8}$$

where, $L$ is the loss function measuring the discrepancy between true target $y_i$ and prediction $F(x_i)$, $N$ is the number of training samples.

**LSTM**

Long Short-Term Memory (LSTM) is a recurrent neural network designed to model sequential data while mitigating the vanishing gradient problem. It maintains a cell state regulated by gates [26].

LSTM equations for time step $t$:

$$i_t=\sigma(U_i h_{t-1}+W_i x_t+b_i) \tag{9}$$

$$f_t=\sigma(U_f h_{t-1}+W_f x_t+b_f) \tag{10}$$

$$o_t=W_o x_{t-1}+U_o x_t+b_o \tag{11}$$

$$\tilde{C}_t=\tanh(U_c x_t+W_c h_{t-1}+b_c) \tag{12}$$

$$C_t=f_t \odot C_{t-1}-i_t \odot \tilde{C}_t \tag{13}$$

$$h_t=o_t \odot \tanh(C_t) \tag{14}$$

where $U_i$ and $W_i$, $U_f$ and $W_f$, $U_o$ and $W_o$, and $U_c$ and $W_c$ are the weights of the input gate, forget gate, output gate, and memory cell, respectively. $b_i$, $b_f$, $b_o$, and $b_c$ are the biases of the input gate, forget gate, output gate, and memory cell, respectively. The intermediate parameter state used for updating is $\tilde{C}_t$. The sigmoid activation and hyperbolic tangent functions are denoted by $\sigma$ and $tanh$, respectively, and $\odot$ denotes the dot product.

**LightGBM**

Light Gradient Boosting Machine (LightGBM) is a high-performance implementation of gradient boosting that uses histogram-based split finding, leaf-wise tree growth, and techniques like GOSS and EFB to speed up training on large-scale data [24].

Additive model update:

$$F_m(x)=F_{m-1}(x)+\eta h_m(x) \tag{15}$$

where, $\eta$ is the learning rate (shrinkage factor), $h_m(x)$ is the decision tree added at iteration $m$.

Regularized objective:

$$Obj = \sum_{i=1}^{N} L(y_i, F(x_i)) + \sum_{m=1}^{M} \Omega(h_m) \tag{16}$$

where, $\Omega(h_m)$ is the regularization term penalizing the complexity of tree $h_m$, $M$ is the total number of trees, $N$ represents the total number of training samples.

**SVR**

Support Vector Regression (SVR) extends support vector machines to regression by fitting a function that is as flat as possible while allowing an ε-insensitive tube around the targets.

Primal optimization problem (ε-SVR)[22]:

$$\min_{w, b, \xi, \xi^*} \|w\|^2 + C \sum_{i=1}^{N} (\xi_i + \xi_i^*) \tag{17}$$

$$y_i - (w^T \phi(x_i) + b) \leq \varepsilon + \xi_i \tag{18}$$

$$(w^T \phi(x_i) + b) - y_i \leq \varepsilon + \xi_i^* \tag{19}$$

$$\xi_i^* \geq 0, \xi_i^* \geq 0 \tag{20}$$

Where, $w$ is the weight vector in feature space; $b$ is the bias term; $\phi(x_i)$ is the feature map (implicitly defined by a kernel); $\varepsilon$ is the width of the insensitive tube (errors inside incur no loss); $\xi_i$ and $\xi_i^*$ are the slack variables for points outside the tube; $C$ is the regularization parameter controlling the trade-off between flatness and tolerance of deviations.

Dual (kernel) form of the predictor:

$$f(x) = \sum_{i=1}^{N} (\alpha_i - \alpha_i^*) K(x_i, x) + b \tag{21}$$

Where, are the lagrange multipliers, $K(x_i, x) = \phi(x_i)^T \phi(x_i)$ is the kernel function.

## 3. 2. 1 Training Process

After data preprocessing and splitting, each of the six prediction models is trained independently. All models use the same training feature matrix and target variables (price and demand are modelled separately). Each model learns the mapping from inputs to outputs on the training set by minimizing a defined loss function. Different models employ their respective training algorithms (e.g., gradient boosting iterations for tree-based models, backpropagation for neural networks, quadratic programming for SVR). All models are trained in a unified computational environment, and after training, the model parameters together with the corresponding normalization parameters are saved for test set prediction.

## 3. 2. 2 Evaluation Metrics

A set of regression metrics is used to quantify how well each model performs, which included regression accuracy, MAE, MAPE, MSE, and coefficient of determination ($R^2$) [41]. These metrics collectively address prediction bias, explanatory power, and practical dependability, thus which supply a quantitative foundation for fair model comparisons.

Suppose the dataset contains $n$ samples in total. For the $i$-th sample, the true value is $y_i$, the predicted value is $\hat{y}_i$, and the mean of the true values is $\overline{y}_i$.

**MAE:**

$$MAE = \frac{1}{n}\sum_{i=1}^{n}|y_i - \hat{y}_i| \qquad (22)$$

**MAPE:**

$$MAE = \frac{100}{n}\sum_{i=1}^{n}\left|\frac{y_i - \hat{y}_i}{y_i}\right| \qquad (23)$$

**MSE:**

$$MPE = \frac{1}{n}\sum_{i=1}^{n}(y_i - \hat{y}_i)^2 \qquad (24)$$

**R²:**

$$R^2 = 1 - \frac{\sum_{i=1}^{n}(y_i - \hat{y}_i)^2}{\sum_{i=1}^{n}(y_i - \overline{y}_i)^2} \qquad (25)$$

# 4. Results and Discussion

This section presents and compares the forecasting performance of the six models (AWMLSTM, CatBoost, GBRT, LSTM, LightGBM, and SVR) on both electricity price and demand. First, the predicted curves and the actual curves are visually analysed to assess each model's ability to track temporal trends, capture peaks and valleys, and reproduce waveform characteristics. Second, a detailed error analysis is provided, including error distribution plots, accuracy tables for relative error thresholds (5% and 10%), and a comprehensive summary of MSE, MAE, $R^2$, and MAPE.

## 4.1 Price and Demand Predictions

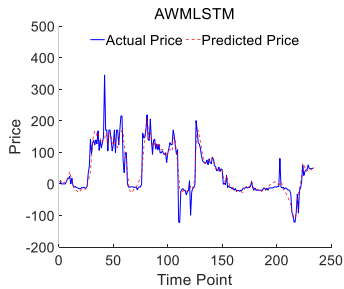
(a) AWMLSTM

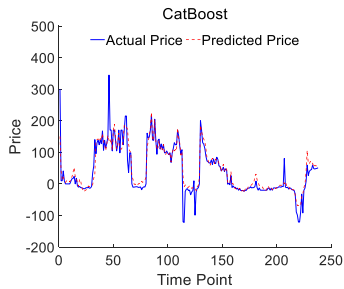
(b) CatBoost

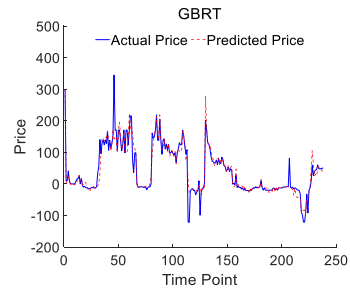
(c) GBRT

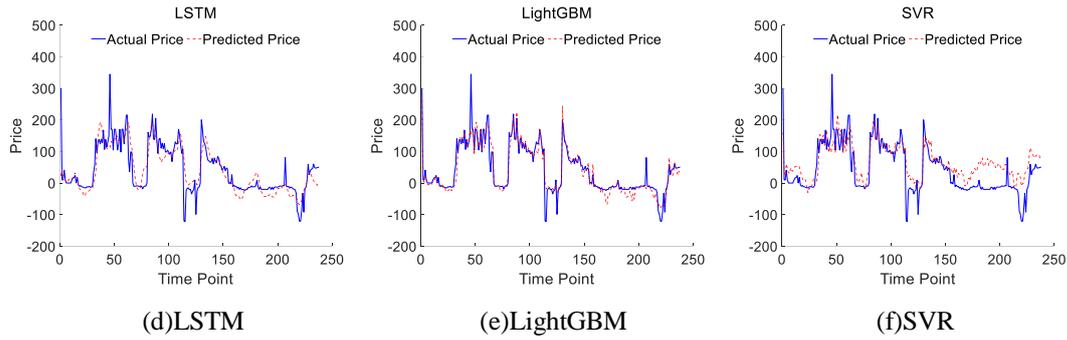

(d)LSTM                 (e)LightGBM              (f)SVR

**Figure 2.** Price prediction

As shown in **Figure 2**, the prediction curves of the predicted electricity prices and actual electricity prices for the six algorithms, AWMLSTM, CatBoost, GBRT, LSTM, LightGBM, and SVR. From the analysis of price trends, the real price signal has a periodic characteristic that is not obvious. For example, the two large cycles around time point 50 and time point 100 are similar, but their detailed variations are quite different. The AWMLSTM, CatBoost, and GBRT models can predict and track the electricity price well. Their predicted time nodes are basically consistent with the real ones, but the prediction accuracy at peaks and valleys is low. The LightGBM model has a weaker prediction capability and shows obvious redundant valleys. The LSTM and SVR models can basically track the price trend, and their predicted time nodes correspond roughly to the real ones, but they have large deviations and distortions.

In **Figure 2(a)**, the predicted electricity price trend of the AWMLSTM model is relatively close to the measured overall trend with roughly aligned peak and valley times. Note that the first several test points are not predicted because the model requires a 24-step historical window to form each input sequence; thus, the prediction starts only after the initial window length. However, the fitting for some sudden peaks is insufficient and the amplitude at sudden valleys is excessively underestimated, which leads to obvious waveform distortion at valleys and a miss of a few peaks and most valleys. AWMLSTM uses wavelet decomposition and an attention mechanism, which however has limited reconstruction of high frequency sudden components such that the attention is dominated by large peaks and causes insufficient peak fitting, while the smoothing characteristic of LSTM and the dominance of low frequency components cause excessive underestimation of valley amplitudes and waveform distortion, which finally results in most valleys and some peaks being missed.

In **Figure 2(b)**, the predicted electricity price trend from the CatBoost model roughly matches the measured overall trend with approximately synchronized peak and valley time positions, but the amplitude deviation of large peaks and valleys is large and the valley waveform has slight distortion, which for example turns an actual sharp peak into a predicted rounded top and misses large peaks and valleys without obvious redundancy. CatBoost uses a symmetric tree structure that forces uniform splits, which makes it difficult to fit local steep changes, and its ordered boosting method gives conservative estimates for outliers, which smooths sharp peaks into rounded tops and distorts valleys, thereby missing large peaks and valleys without redundancy.

In **Figure 2(c)**, GBRT basically reproduces the trend of the real electricity price with almost identical peak and valley time positions, but there are large residuals in segments where electricity price suddenly rises or falls sharply, and the peak and valley values have slight overshoot or undershoot, which suppresses the actual features at large amplitude peaks and valleys and misses large peaks and valleys without obvious redundancy. GBRT uses iterative residual fitting with shallow trees that have a lag in fitting sudden segments, which leads to overshoot or undershoot, and its loss function suppresses extreme amplitudes such that the features of large peaks and valleys are weakened, resulting in obvious missed peaks and valleys without redundancy.

In **Figure 2(d)**, the predicted electricity price trend from the LSTM model deviates greatly from the measured overall trend with obvious shifts in peak and valley time positions and a significant amplitude deviation, where the overall fluctuation width is compressed to half of the actual width and the waveform shape is severely distorted, such that a relatively flat U shaped valley in reality is predicted as a steep V shaped valley and the model clearly under predicts fine details, which misses most peaks and valleys. LSTM is sensitive to rapid fluctuations, and improper hyperparameters can easily cause gradient drift that leads to overall trend deviation, while its memory unit smooths short term sudden changes and compresses the amplitude to half, and its response lag changes a flat U valley into a V valley, whereas the forget gate discards details, which causes most peaks and valleys to be omitted.

In **Figure 2(e)**, the predicted electricity price trend from the LightGBM model is roughly consistent with the measured overall trend with moderate fitting of peak and valley time positions, but the amplitude deviation is large such that large fluctuation peaks are slightly underestimated and the valley prediction error is large, and the waveform shape has distortion which turns a relatively flat U shaped valley in reality into a predicted steep V shaped valley with obvious redundant valleys. LightGBM uses GOSS that preferentially retains large gradient samples for peaks and randomly downsamples small gradient samples for valleys, which leads to the prediction of a flat U valley as a steep V valley, and its feature bundling may produce redundant valleys, so peaks are slightly underestimated and valley errors are large.

In **Figure 2(f)**, the predicted electricity price trend from the SVR model deviates greatly from the measured overall trend with severely misaligned peak and valley time positions and obviously abnormal peaks and valleys, where the amplitude deviation is significant and the overall fluctuation width is compressed to half of the actual width, and the waveform shape is severely distorted such that a steep V shaped valley in reality is predicted as a wide U shaped valley or vice versa, which misses multiple actual peaks and valleys and produces several false fluctuations. SVR has no explicit temporal dependence and models each time point independently, which causes severe misalignment of peaks and valleys, and improper settings of the kernel function and parameters C and epsilon compress the amplitude to half and generate false fluctuations or distortions that swap V shape and U shape, thereby causing both missing and false reports.

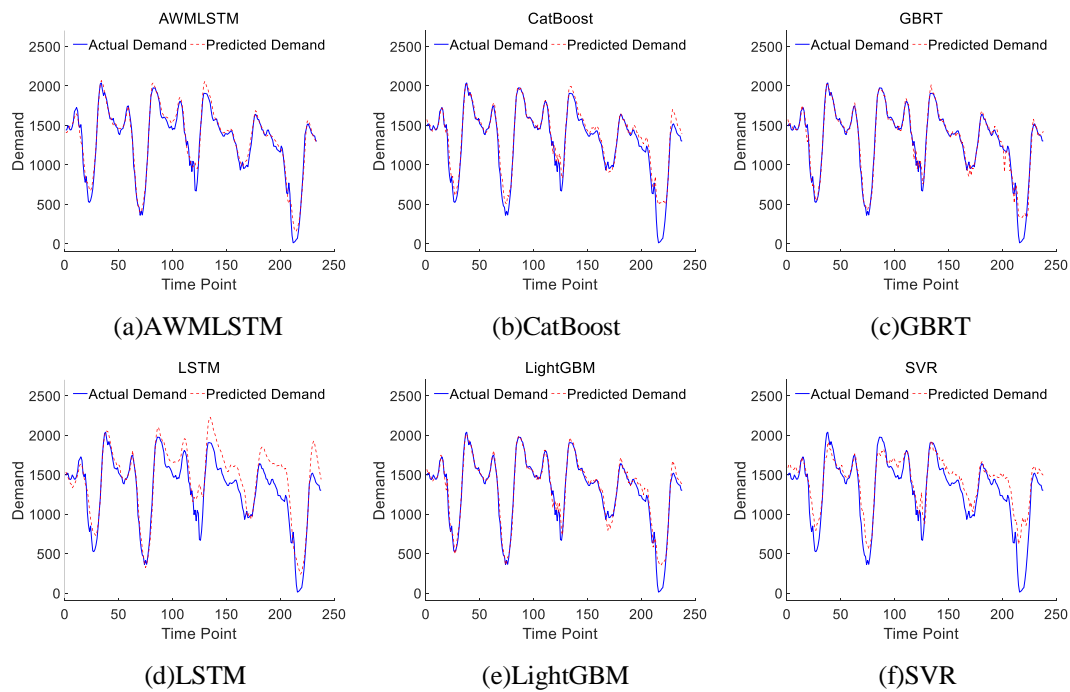

(a)AWMLSTM  (b)CatBoost  (c)GBRT

(d)LSTM  (e)LightGBM  (f)SVR

**Figure 3.** Demand prediction

As shown in **Figure 3**, the prediction curves of the predicted electricity demand and actual electricity demand for the six algorithms, AWMLSTM, CatBoost, GBRT, LSTM, LightGBM, and SVR. From the analysis of the demand forecasting trends, it can be seen that the actual demand signal exhibits periodic like characteristics. The cycle lengths of all models are basically consistent, and the time positions of peaks and troughs are also basically consistent. The four models, AWMLSTM, CatBoost, GBRT, and LightGBM, achieve good prediction and tracking of electricity demand, but their prediction accuracy at peaks and troughs is relatively low. Among them, GBRT and LightGBM show a few redundant troughs. The LSTM and SVR models can basically track the demand trend, but they suffer from significant deviations and distortions.

The main reason for low accuracy at peaks and troughs is that extreme demand events often associate with rare or irregular external factors, such as sudden temperature shifts or holidays. Thus, these samples appear infrequently in historical data. For tree based models like GBRT and LightGBM, their decision boundaries fit common patterns well but lack enough support at extremes, which creates a few redundant troughs. For sequential models like LSTM and AWMLSTM, a natural lag exists between input and output, and their internal state tends to smooth abrupt changes; as a result, sharp peaks and troughs become less precise. The SVR model has no explicit time dependence, so it treats rapid demand changes as noise and produces larger deviations. Although AWMLSTM and CatBoost use attention or ordered boosting to improve robustness, their structure still cannot perfectly capture the most extreme points.

## 4.2 Error Analysis

The results for error are presented separately for price prediction Section 4.2.1, demand prediction Section 4.2.2, and an overall comparison across both tasks Section 4.2.3.

## 4.2.1 Error of Price Prediction

The price error curve characteristics of the six algorithms, AWMLSTM, CatBoost, GBRT, LSTM, LightGBM, and SVR, are shown in **Figure 4**, and the corresponding errors at 5% and 10% are shown in **Table 2**. The four models AWMLSTM, CatBoost, GBRT and LightGBM perform relatively well in electricity price forecasting. LSTM and SVR exhibit larger prediction deviations. In addition, the predictions of CatBoost and SVR tend to be systematically overestimated, whereas those of LightGBM are systematically underestimated.

Quantitative results from **Table 2** reveal significant differences among the models in the percentage of samples with relative errors within 5% and 10%. Among all models, GBRT achieves the best performance. It has 21.74% of samples with errors within 5% and 34.78% within 10%. CatBoost ranks second with corresponding proportions of 18.70% and 33.04%. AWMLSTM lies in the middle range with 16.81% and 26.99%. LightGBM achieves 13.91% within 5% and 26.52% within 10%. These figures slightly outperform SVR with 6.96% and 17.39% and LSTM with 6.52% and 12.17%. The latter two models show clearly insufficient predictive ability in the high precision range, namely within 5% error. LSTM performs worst among all models.

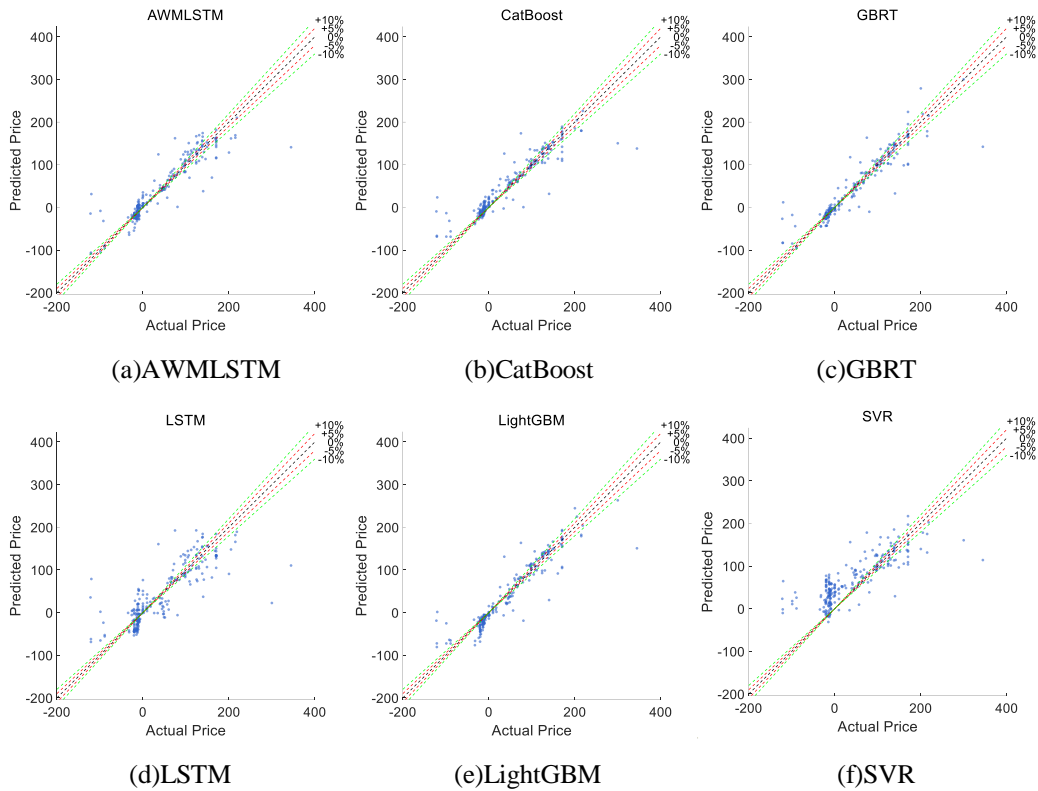

**Figure 4.** Price Error Analysis

**Table 2. Accuracy prediction demand errors of the algorithms**

| Accuracy | AWMLSTM | CatBoost | GBRT | LSTM | LightGBM | SVR |
|---|---|---|---|---|---|---|
| ±5% | 16.81% | 18.70% | 21.74% | 6.52% | 13.91% | 6.96% |
| ±10% | 26.99% | 33.04% | 34.78% | 12.17% | 26.52% | 17.39% |

Overall, tree-based models such as GBRT, CatBoost and LightGBM generally outperform LSTM and SVR. However, even the best model GBRT still has more than 65% of samples with a relative error exceeding 10%. This fact indicates that electricity price forecasting remains a challenging task.

These results reflect the complexity of electricity price forecasting and the differences in fitting capability and bias characteristics among the models. One reason for the poor performance of LSTM and SVR is the strong nonlinearity and high volatility of electricity price series. LSTM lacks sufficient adaptability to short term abrupt changes. SVR faces difficulties in kernel function selection and hyperparameter tuning under high dimensional features, and this often leads to systematic overestimation.

Another observation is that the overestimation in CatBoost and SVR and the underestimation in LightGBM indicate different bias directions when these models learn the tails of the price distribution, especially extreme high and low prices. CatBoost tends to be conservative in fitting high price regions or gets pulled by outliers. LightGBM may shift its predictions downward due to its regularization or leaf growth strategy.

A further point is that even the best model GBRT still has about two thirds of samples with errors exceeding 10%. This outcome is not a defect of the model itself. Instead it results from the inherently difficult to predict nature of electricity prices, which multiple uncertain factors influence. These factors include weather, load, renewable energy output and market behaviour. Therefore practical applications can integrate multiple complementary models, for example combine GBRT with AWMLSTM, or introduce error correction mechanisms. These approaches can further improve forecasting accuracy.

### 4.2.2 Error of Demand Prediction

The demand error curve characteristics of the six algorithms, AWMLSTM, CatBoost, GBRT, LSTM, LightGBM, and SVR, are shown in **Figure 5**, and the corresponding errors at 5% and 10% are shown in **Table 3**. It can be seen that AWMLSTM, GBRT, and LightGBM provide relatively good demand forecasts, while the predictions of LSTM and SVR deviate significantly. In addition, CatBoost, LSTM, and SVR all tend to systematically overestimate actual demand.

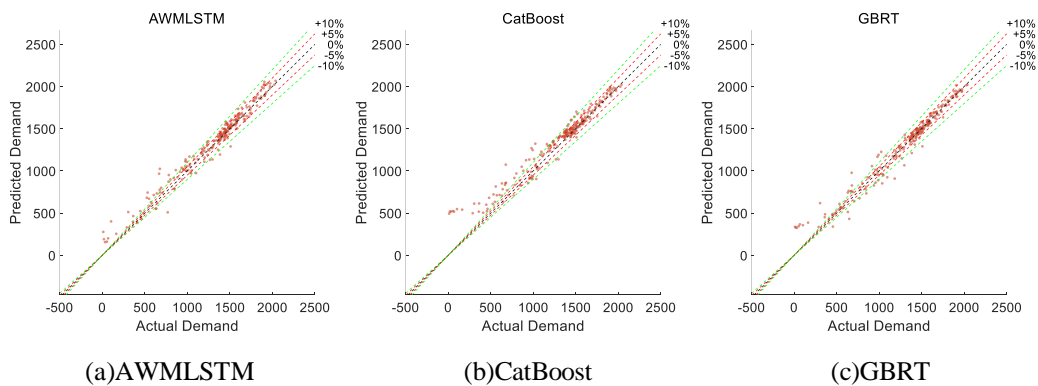

(a)AWMLSTM    (b)CatBoost    (c)GBRT

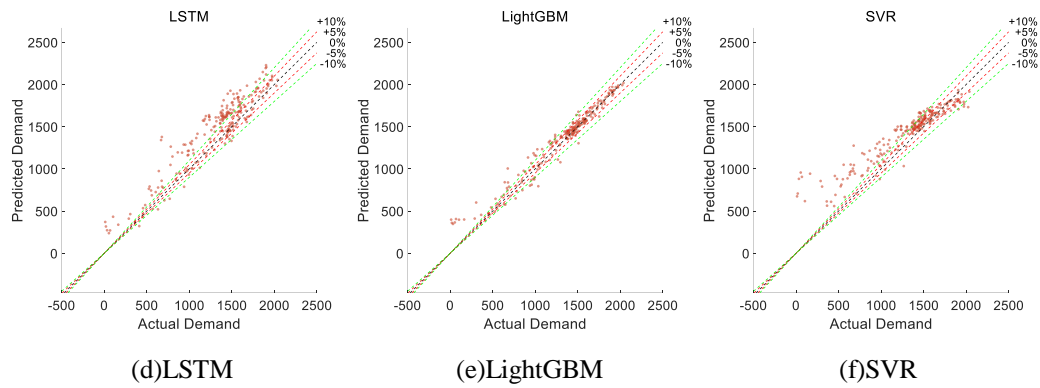

(d)LSTM                (e)LightGBM              (f)SVR

**Figure 5.** Demand Error Analysis

**Table 3. Accuracy prediction demand errors of the algorithms**

| Accuracy | AWMLSTM | CatBoost | GBRT | LSTM | LightGBM | SVR |
|---|---|---|---|---|---|---|
| ±5% | 64.53% | 59.24% | 74.37% | 29.41% | 66.81% | 31.09% |
| ±10% | 83.76% | 78.15% | 84.87% | 45.38% | 81.51% | 60.50% |

In the demand forecasting task, the prediction accuracies of the six models exhibit substantial variation, where accuracy is defined as the proportion of samples with absolute errors within 5% and within 10%, respectively. Among all models, GBRT achieves the highest accuracy, with 74.37%t of samples below the 5% error threshold and 84.87% below the 10% threshold. LightGBM attains 66.81% and 81.51%, whereas AWMLSTM attains 64.53% and 83.76%; both models demonstrate comparable performance and rank next to GBRT. CatBoost, with 59.24% and 78.15%, occupies an intermediate position. In contrast, SVR with 31.09% and 60.50% and LSTM with 29.41% and 45.38% lag markedly behind the aforementioned models, and LSTM shows the lowest accuracy overall.

### 4.2.3 Error of Overall Prediction

A joint examination of the results from **Figure 4** and **Figure 5** confirms that AWMLSTM, GBRT, and LightGBM produce reliable demand forecasts. Specifically, GBRT yields the best prediction precision, as reflected by its superior error distribution. LightGBM and AWMLSTM achieve slightly lower but still favorable accuracies. By contrast, LSTM and SVR not only exhibit large prediction deviations and the worst accuracies but also, together with CatBoost, display a systematic overestimation of actual demand. These findings indicate that traditional tree-based ensemble methods, especially GBRT, outperform the deep-learning model LSTM as well as its variant AWMLSTM, notwithstanding the improvement gained by the attention mechanism in AWMLSTM. Furthermore, the positive bias observed in SVR and CatBoost requires attention in practical applications.

**Table 4. Prediction errors of the algorithms**

| Algorithm | Data | MSE | MAE | $R^2$ | MAPE |
|---|---|---|---|---|---|
| AWMLSTM | Price | 817.07 | 15.96 | 0.86 | >250% |
| | Demand | 6922.54 | 61.09 | 0.96 | 23.3% |
| CatBoost | Price | 859.10 | 15.45 | 0.85 | >250% |
| | Demand | 14785.64 | 79.73 | 0.92 | 49.9% |
| GBRT | Price | 713.25 | 13.25 | 0.88 | >150% |
| | Demand | 7270.12 | 54.41 | 0.96 | 32.0% |
| LSTM | Price | 1901.94 | 28.30 | 0.68 | >200% |
| | Demand | 42168.29 | 163.66 | 0.78 | 39.7% |
| LightGBM | Price | 742.37 | 15.50 | 0.87 | 90.1% |
| | Demand | 8907.93 | 63.09 | 0.95 | 36.3% |
| SVR | Price | 2624.65 | 38.74 | 0.55 | >300% |
| | Demand | 46115.00 | 152.44 | 0.76 | 78.8% |

The MSE, MAE, $R^2$, and MAPE results of price and demand for the six algorithms AWMLSTM, CatBoost, GBRT, LSTM, LightGBM, and SVR are shown in Table 4. In the price prediction task, the mean absolute percentage error MAPE of all models exceeds 90%, a sign of generally poor accuracy. Among all models, GBRT achieves the best relative performance with a coefficient of determination $R^2$ of 0.88 and a mean squared error MSE of 713.25, values significantly lower than those of LSTM and SVR. LightGBM ranks second with an $R^2$ of 0.87 and a MAPE of 90.1%, the only value below 100%. In contrast, the $R^2$ of LSTM drops to 0.68, while SVR has an $R^2$ of only 0.55. Both models show MSE values above 1900, and the MAPE of SVR exceeds 300%. These results suggest that traditional tree based ensemble methods maintain a relative advantage in price prediction. However, no model achieves accurate forecasts, a failure that may stem from high price volatility or insufficient features.

Demand prediction yields much better results than price prediction. Several models achieve an $R^2$ above 0.95. AWMLSTM delivers the best overall performance with an $R^2$ of 0.96, an MSE of 6922.54, the lowest value among all models, and a MAPE of only 23.3%, also the lowest. GBRT performs equally well with an $R^2$ of 0.96 and the lowest MAE of 54.41, while its MSE is slightly higher than that of AWMLSTM. LightGBM follows closely with an $R^2$ of 0.95 and error metrics comparable to the top models. CatBoost sits in the middle with an $R^2$ of 0.92 but a MAPE of 49.9%. LSTM and SVR lag significantly behind. LSTM has an R² of only 0.78 and an MSE above 42000, while SVR achieves an R² of only 0.76 and a MAPE as high as 78.8%. Therefore, AWMLSTM and GBRT are the most recommended models for demand prediction, with LightGBM as a reliable alternative. LSTM and SVR are unsuitable for this task.

A cross task comparison reveals several main findings. Demand prediction is substantially easier than price prediction. Every model shows a higher R² for demand data than for price data, and the MAPE values for demand are much lower than those for price. This difference indicates that demand series possess stronger predictability. GBRT performs consistently well in both tasks and is the most generalizable model. AWMLSTM outperforms GBRT in demand prediction but falls slightly behind in price prediction. LightGBM ranks among the top tier in both tasks. LSTM and SVR perform poorly in both tasks, with SVR almost always the worst across all metrics. This poor performance proves that these two model classes are not suitable for price or demand prediction on this dataset. Furthermore, the extremely high MAPE values in price prediction suggest that this task suffers from data noise, insufficient features, or inherent randomness.

## 5. Conclusion

This study compares six algorithms for short term electricity price forecast and demand forecast in the South Australian electricity market. The results show that tree-based models, especially GBRT, perform better than LSTM and SVR for price prediction. However, all models struggle with high price volatility. For demand prediction, AWMLSTM and GBRT achieve relatively higher accuracy. The following conclusions can be drawn from this study:

1) In price prediction, AWMLSTM, CatBoost, and GBRT track overall trends with aligned peaks and valleys, but AWMLSTM underestimates valleys, CatBoost fails on steep changes, GBRT suffers from lag-induced overshoot, and LightGBM shows redundant valleys.

2) LSTM and SVR perform not relatively accurate in price prediction. Their fluctuation width halves, time nodes misalign, and severe waveform distortions appear.

3) Tree-based models outperform LSTM and SVR in price prediction, and GBRT achieves the best R² of 0.88. Yet more than 65% of its predictions have errors above 10%. CatBoost and SVR overestimate extremes, whereas LightGBM underestimates them.

4) Demand has clear periodicity, and all models capture its cycle timing. AWMLSTM, CatBoost, GBRT, and LightGBM track demand well but have low accuracy at peaks and troughs. GBRT and LightGBM show redundant troughs because of rare extreme samples, while LSTM and SVR exhibit large deviations and systematic overestimation.

5) Demand prediction proves easier than price prediction because AWMLSTM and GBRT achieve an $R^2$ of 0.96 and a MAPE below 32%. AWMLSTM, GBRT, and LightGBM provide reliable demand predictions, and GBRT is best. For GBRT, 74.37% of samples fall within 5% error and 84.87% within 10% error.

6) LSTM and SVR perform poorly in both price and demand prediction tasks, whereas GBRT is the most generalizable model across tasks.

The future study should improve price forecast through hybrid models such as tree plus transformer to capture high frequency spikes, data augmentation for extreme events, and error correction for bias. For demand prediction, refinement of peak and trough accuracy requires the incorporation of external factors like weather and holidays. Adaptive models that switch strategies based on volatility regimes can yield more robust energy forecasts for both price and demand tasks.

**References**

1. Kirschen, D.S. and G. Strbac, *Fundamentals of power system economics*. 2018: John Wiley & Sons.

2. Parker, G.G., B. Tan, and O. Kazan, *Electric power industry: Operational and public policy challenges and opportunities.* Production and Operations Management, 2019. **28**(11): p. 2738-2777.

3. Weron, R., *Electricity price forecasting: A review of the state-of-the-art with a look into the future.* International journal of forecasting, 2014. **30**(4): p. 1030-1081.

4. Hong, T., et al., *Energy forecasting: A review and outlook.* IEEE Open Access Journal of Power and Energy, 2020. **7**: p. 376-388.

5. Nowotarski, J. and R. Weron, *Recent advances in electricity price forecasting: A review of probabilistic forecasting.* Renewable and Sustainable Energy Reviews, 2018. **81**: p. 1548-1568.

6. Contreras, J., et al., *ARIMA models to predict next-day electricity prices.* IEEE transactions on power systems, 2003. **18**(3): p. 1014-1020.

7. Lago, J., F. De Ridder, and B. De Schutter, *Forecasting spot electricity prices: Deep learning approaches and empirical comparison of traditional algorithms.* Applied Energy, 2018. **221**: p. 386-405.

8. Lago, J., et al., *Forecasting day-ahead electricity prices: A review of state-of-the-art algorithms, best practices and an open-access benchmark.* Applied Energy, 2021. **293**: p. 116983.

9. Brusaferri, A., et al., *Bayesian deep learning based method for probabilistic forecast of day-ahead electricity prices.* Applied Energy, 2019. **250**: p. 1158-1175.

10. Cerqueira, V., L. Torgo, and I. Mozetič, *Evaluating time series forecasting models: An empirical study on performance estimation methods.* Machine Learning, 2020. **109**(11): p. 1997-2028.

11. Rai, A. and O. Nunn, *On the impact of increasing penetration of variable renewables on electricity spot price extremes in Australia.* Economic analysis and policy, 2020. **67**: p. 67-86.

12. Yan, G. and L. Han, *The impact of rooftop solar on wholesale electricity demand in the Australian National Electricity Market.* Frontiers in Energy Research, 2023. **11**: p. 1197504.

13. Cutler, N.J., et al., *High penetration wind generation impacts on spot prices in the Australian national electricity market.* Energy Policy, 2011. **39**(10): p. 5939-5949.

14. Forrest, S. and I. MacGill, *Assessing the impact of wind generation on wholesale prices and generator dispatch in the Australian National Electricity Market.* Energy policy, 2013. **59**: p. 120-132.

15. Cornell, C., N.T. Dinh, and S.A. Pourmousavi, *A probabilistic forecast methodology for volatile electricity prices in the Australian National Electricity Market.* International Journal of Forecasting, 2024. **40**(4): p. 1421-1437.

16. Csereklyei, Z., S. Qu, and T. Ancev, *The effect of wind and solar power generation on wholesale electricity prices in Australia.* Energy Policy, 2019. **131**: p. 358-369.

17. Mwampashi, M.M., C.S. Nikitopoulos, and A. Rai, *From 30-to 5-minute settlement rule in the NEM: An early evaluation.* Energy Policy, 2024. **194**: p. 114305.

18. Gonçalves, R. and F. Menezes, *The price impacts of the exit of the Hazelwood coal power plant.* Energy Economics, 2022. **116**: p. 106398.

19. O'Connor, C., et al., *A review of electricity price forecasting models in the day-ahead, intra-day, and balancing markets.* Energies, 2025. **18**(12): p. 3097.

20. Makridakis, S., E. Spiliotis, and V. Assimakopoulos, *Statistical and Machine Learning forecasting methods: Concerns and ways forward.* PloS one, 2018. **13**(3): p. e0194889.

21. Uniejewski, B., R. Weron, and F. Ziel, *Variance stabilizing transformations for electricity spot price forecasting.* IEEE Transactions on Power Systems, 2017. **33**(2): p. 2219-2229.

22. Smola, A.J. and B. Schölkopf, *A tutorial on support vector regression.* Statistics and computing, 2004. **14**(3): p. 199-222.

23. Friedman, J.H., *Greedy function approximation: a gradient boosting machine.* Annals of statistics, 2001: p. 1189-1232.

24. Ke, G., et al., *Lightgbm: A highly efficient gradient boosting decision tree.* Advances in neural information processing systems, 2017. **30**.

25. Prokhorenkova, L., et al., *CatBoost: unbiased boosting with categorical features.* Advances in neural information processing systems, 2018. **31**.

26. Hochreiter, S. and J. Schmidhuber, *Long short-term memory.* Neural computation, 1997. **9**(8): p. 1735-1780.

27. Yuan, X., et al., *Deep learning with spatiotemporal attention-based LSTM for industrial soft sensor model development.* IEEE Transactions on Industrial Electronics, 2020. **68**(5): p. 4404-4414.

28. Ghimire, S., et al., *Two-step deep learning framework with error compensation technique for short-term, half-hourly electricity price forecasting.* Applied Energy, 2024. **353**: p. 122059.

29. Wang, D., et al., *Electricity Price Instability over Time: Time Series Analysis and Forecasting.* Sustainability, 2022. **14**(15): p. 9081.

30. Abroun, M., et al., *Predicting long-term electricity prices using modified support vector regression method.* Electrical Engineering, 2024. **106**(4): p. 4103-4114.

31. Hu, J., et al., *A data driven model based approach for medium-to-long-term electricity price forecasting in power markets.* Scientific Reports, 2025. **15**(1): p. 37046.

32. Das, A., S. Schlüter, and L. Schneider, *Electricity Price Prediction Using Multikernel Gaussian Process Regression Combined With Kernel-Based Support Vector Regression.* Journal of Forecasting, 2026.

33. Kuşkaya, S. and F. Bilgili, *Forecasting electricity price index with machine learning models and strategies.* Quality & Quantity, 2026. **60**(1): p. 2651-2678.

34. Nasios, I. and K. Vogklis, *Blending gradient boosted trees and neural networks for point and probabilistic forecasting of hierarchical time series.* International Journal of Forecasting, 2022. **38**(4): p. 1448-1459.

35. Bâra, A., S.-V. Oprea, and A.-C. Băroiu, *Forecasting the Spot Market Electricity Price with a Long Short-Term Memory Model Architecture in a Disruptive Economic and Geopolitical Context.* International Journal of Computational Intelligence Systems, 2023. **16**(1): p. 130.

36. Zi, X., et al., *A Deep Learning Method for Photovoltaic Power Generation Forecasting Based on a Time-Series Dense Encoder.* Energies, 2025. **18**(10): p. 2434.

37. Yang, G., et al., *Short-term Price Forecasting Method in Electricity Spot Markets Based on Attention-LSTM-mTCN.* Journal of Electrical Engineering & Technology, 2022. **17**(2): p. 1009-1018.

38. Marcjasz, G., et al., *Distributional neural networks for electricity price forecasting.* Energy Economics, 2023. **125**: p. 106843.

39. Berrisch, J. and F. Ziel, *Multivariate probabilistic CRPS learning with an application to day-ahead electricity prices.* International Journal of Forecasting, 2024. **40**(4): p. 1568-1586.

40. Qin, Y., et al., *A dual-stage attention-based recurrent neural network for time series prediction.* arXiv preprint arXiv:1704.02971, 2017.

41. Khoshvaght, H., et al., *A critical review on selecting performance evaluation metrics for supervised machine learning models in wastewater quality prediction.* Journal of Environmental Chemical Engineering, 2025. **13**(6): p. 119675.